\documentclass{ecai}  

\usepackage{graphicx}
\usepackage{latexsym}
\usepackage{amssymb}
\usepackage{amsmath}
\usepackage{amsthm}
\usepackage{booktabs}
\usepackage{enumitem}
\usepackage{color}
\usepackage{multirow}
\usepackage{pifont}
\usepackage{adjustbox}
\usepackage{appendix}

\begin{document}

\begin{frontmatter}

\title{EdgeNAT: Transformer for Efficient Edge Detection}

\author[A]{\fnms{Jinghuai}~\snm{Jie}}
\author[A]{\fnms{Yan}~\snm{Guo}\thanks{Corresponding Author. Email: guoyan@ustc.edu.cn.}}
\author[A]{\fnms{Guixing}~\snm{Wu}}
\author[A]{\fnms{Junmin}~\snm{Wu}}
\author[A]{\fnms{Baojian}~\snm{Hua}\thanks{Corresponding Author. Email: bjhua@ustc.edu.cn.}}

\address[A]{University of Science and Technology of China}
%

\begin{abstract}
 Transformers, renowned for their powerful feature extraction capabilities, have played an increasingly prominent role in various vision tasks. Especially, recent advancements present transformer with hierarchical structures such as {\it Dilated Neighborhood Attention Transformer} (DiNAT), demonstrating outstanding ability to efficiently capture both global and local features. However, transformers' application in edge detection has not been fully exploited. In this paper, we propose {\it EdgeNAT}, a one-stage transformer-based edge detector  with DiNAT as the encoder, capable of extracting object boundaries and meaningful edges both accurately and efficiently. On the one hand, EdgeNAT captures global contextual information and detailed local cues with DiNAT, on the other hand, it enhances feature representation with a novel SCAF-MLA decoder by utilizing both inter-spatial and inter-channel relationships of feature maps. Extensive experiments on  multiple datasets show that our method achieves state-of-the-art performance on both RGB and depth images. Notably, on the widely used BSDS500 dataset, our L model achieves impressive performances, with ODS F-measure and OIS F-measure of 86.0\%, 87.6\% for multi-scale input,and 84.9\%, and 86.3\% for single-scale input, surpassing the current state-of-the-art EDTER by 1.2\%, 1.1\%, 1.7\%, and 1.6\%, respectively. Moreover, as for throughput, our approach runs at 20.87 FPS on RTX 4090 GPU with single-scale input. Code: https://github.com/jhjie/EdgeNAT.
\end{abstract}

\end{frontmatter}


\section{Introduction}

\begin{figure}[t]
    \centering
    \includegraphics[width=0.9\linewidth]{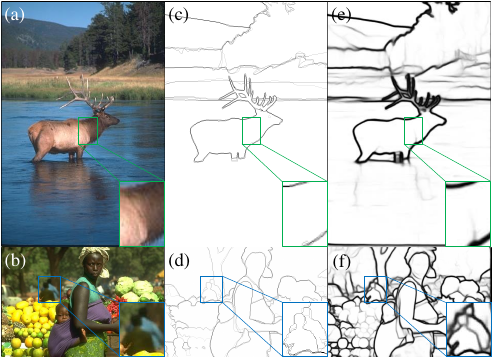}
    \caption{Illustration of the predictions of our EdgeNAT. (a, b): Input images from BSDS500. (c, d): the corresponding groundtruth. (e, f): Detected edges by EdgeNAT. (e) shows EdgeNAT doesn't extracts an edge in the neck area of the deer on the presence intense color variation.(f) shows EdgeNAT accurately extracts edges of distant and blurred objects. }
    \label{fig:fig1}
\end{figure}

Edge detection is fundamental for various computer vision tasks\cite{2,3,4}. The primary objective of edge detection is to precisely extract object boundaries and visually salient edges from input images. As illustrated in Figure \ref{fig:fig1}, inherent challenges of this task include the presence of distant objects, blurring boundary in complex backgrounds, intense color variations within an objects, etc. Therefore, it requires appropriate representation of not only local features like color and texture, but also global semantic information to suppress noise as well as to distinguish object boundary from complex background. 

Traditional edge extraction methods\cite{5,6} mostly rely on local  information, such as variation of color and texture. CNN-based deep learning based edge detectors can learn global and semantic features\cite{7,8} with expansion of receptive field, but are likely to lose detail information. To preserve both intricate local information as well as global context, former deep learning detectors \cite{9,10} employ multi-level aggregation to effectively integrate global features and local details. To mitigate the limitations in the absence of a hierarchical structure in ViT\cite{12}, the first transformer based edge detector EDTER\cite{11} implements a two-stage approach to obtain and combine global features and local details, which demonstrates superior edge detection ability than CNN-based detectors. However, the computational burden known for vision transformer is exacerbated by EDTER's two-stage design.

\begin{figure*}[t]
  \centering
  \includegraphics[width=\textwidth]{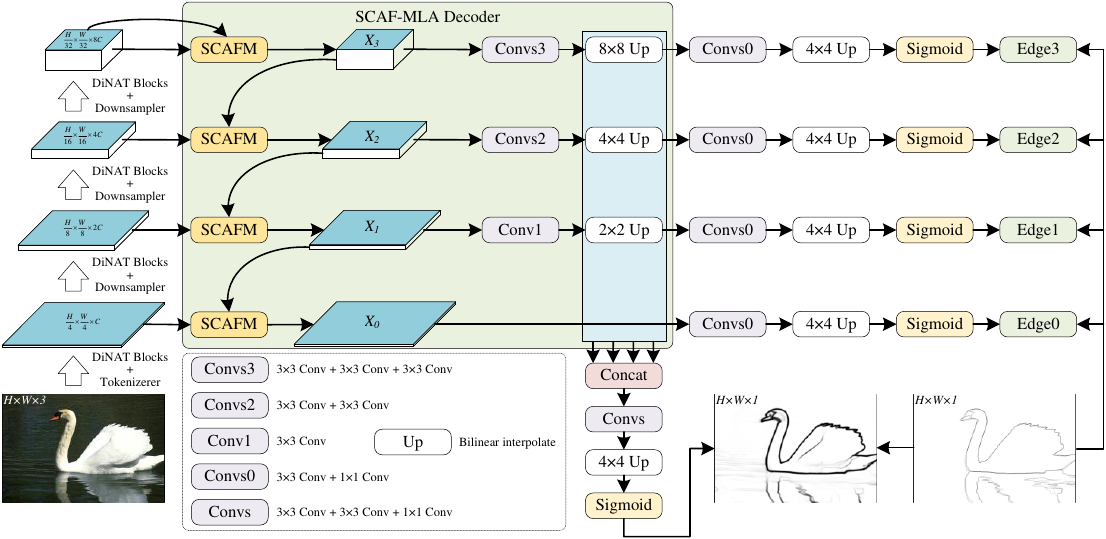}
  \caption{The overall framework of our proposed EdgeNAT. SCAFM is illustrated in Figure \ref{fig:fig3}.}
  \label{fig:overview}
\end{figure*}

Recently, DiNAT\cite{15}, an improved hierarchical transformer combining both neighbor attention and dilated neighbor attention, has exemplified significant progress in various vision tasks. 
Since DiNAT is able to preserve locality, maintain translation equivariance, expand the receptive field exponentially, and capture longer-range inter-dependencies, edge detector based on it (Figure \ref{fig:overview}) could abandon the two-stage design, significantly improving throughput. Furthermore, to take better usage of rich channels information in the feature map generated by transformer-based encoder, we introduce a novel decoder, Spatial and Channel Attention Fusion-Multi-Level Aggregation (SCAF-MLA). The Spatial and Channel Attention Fusion Module (SCAFM) of the decoder integrates both spatial and channel attention concurrently. As a whole, our detector is capable of extracting local detail information at lower levels, which is beneficial to the detection of edges associated with distant blurred objects, and extracting global semantic feature information at higher levels, which is beneficial to mitigating excessive noise within the object and to distinguishing inconspicuous edges. 

With elaborate design, our models exhibit excellent capability of generating accurate and crisp edge maps.
To verify the scalability of our edge detector, we further propose five versions of models with varying sizes, following DiNAT's configuration. Our contributions in this paper can be summarized as follows:
    (1) We introduce EdgeNAT, a one-stage Transformer-based edge detector, which enables local and global features extraction, leading to speedy and precise edge detection.
    (2) We propose an innovative feature fusion module, SCAFM, to enhance the feature representation generated by the encoder. We further design the SCAF-MLA decoder based on it.
    (3) Extensive experiments conducted on widely recognized edge detection benchmarks, such as BSDS500 and NYUDv2, demonstrate the superior performance and high efficiency of our model when compared to state-of-the-art methods. Adaptability and flexibility of our architecture are also verified on five variants of our model.

\section{Related Work}

\textbf{Edge Detection.}
Early edge detectors\cite{5,6} mainly rely on local features, like significant variation of color, texture and intensity to detect edges. Machine learning-based methods\cite{18,19} employ hand-crafted low-level features to train classifiers and achieve impressive performance compared to earlier approaches. Such methods are always ignorant of global information and semantic boundaries. CNN-based deep learning techniques are able to expand receptive fields to capture global features, thus yield remarkable progress in edge detection. DeepEdge\cite{7} employs a multi-scale CNN to classify edge candidate points extracted by Canny edge detector. Recent methods have further enhanced edge detection by exploiting hierarchical and multi-scale feature maps CNN encoders produce. \cite{10,9} learn rich hierarchical features by supervising the layers at each level, leading to improved detection performance. BDCN\cite{23}, on the other hand, achieves greater accuracy through a bidirectional feature processing structure. PiDiNet\cite{24} introduces pixel-differential convolutional integration into the CNN model. EDTER\cite{11} is the first attempt to introduce Vision Transformer (ViT)\cite{12} for edge detection tasks. To capture multi-scale features, EDTER proposes a two-stage architecture to remedy the lack of hierarchical structure in ViT. The first stage focuses on global feature while the second stage focuses on local features. Features learned in both stages are fused, resulting in significant improvements in performance and achieving SOTA in edge detection task. PEdger\cite{27} enhances edge detection performance by leveraging information obtained from different training moments and heterogeneous structures. UAED\cite{28} investigates the subjectivity and ambiguity of different annotations through uncertainty based on the fact that dataset labels have multiple annotations.

\noindent
\textbf{Vision Transformer.}
Since the introduction of ViT\cite{12}, transformers have been widely used in vision field\cite{29,30,31}. After years of development, Transformer with multi-scale hierarchical structure are playing increasingly important role in  downstream vision tasks. Swin Transformer\cite{13} proposes Window Self Attention (WSA) and Shift Window Self Attention (SWSA), with SWSA expanding the receptive field, enabling it to capture both local and global features. NAT\cite{14} proposes Neighborhood Attention (NA), the first efficient and scalable sliding window attention mechanism, which restricts self-attention to localised windows and preserves translation equivariance. DiNAT\cite{15} extends NA to Dilated Neighborhood Attention (DiNA), which expands receptive fields exponentially and thus captures long-range inter dependency and global features. Besides, Neighborhood Attention Extension (NATTEN)\cite{14} is developed to better implement NA and DiNA as an extension to PyTorch with an efficient CUDA kernel.

\noindent
\textbf{Feature Fusion Module.} The feature fusion module is commonly used in edge detection and other vision tasks to strengthen feature representations, which is crucial to improving the accuracy. 
SENet\cite{32} investigates channel relationships and introduces a novel architectural unit, the Squeeze-and-Excitation (SE) block, enhances global feature extraction by computing channel attention using global average pooling. 
CBAM\cite{33} employs both global average pooling and global maximum pooling to compute attention maps on two separate dimensions, namely, channel attention and spatial attention, the latter being overlooked by SENet. 
CBAM is able to extract informative features by blending cross-channel and spatial information together. 
ECA\cite{48} proposes a local cross channel interaction strategy implemented via 1D convolution and a method to adaptively select kernel size of 1D convolution.  
PP-LiteSeg\cite{34} introduces UAFM, a feature fusion module that leverages channel attention or spatial attention to enrich the representation of fused features, with spatial and channel attention modules exploiting inter-spatial and inter-channel relationships of the input features. 

\section{EdgeNAT}

Figure \ref{fig:overview} illustrates the overall framework of EdgeNAT, a one-stage end-to-end edge detector. DiNAT is employed  as the encoder since it exhibits exceptional performance in preserving locality, maintaining translation equivariance, expanding receptive field, and capturing long-range dependencies,etc. SCAF-MLA, a novel decoder with SCAFM to exploit both spatial and channel features from feature maps, is introduced to effectively facilitate feature fusion. We further improve the performance of  SCAF-MLA by pre-fusion, that is, for the fusing operation, the feature channels of each layer are reduced to the number of channels in first level of the encoder, denoted as C in Figure \ref{fig:overview}, rather than to 1.

\subsection{Review Dilated Neighborhood Attention Transformer}

Below is a brief introduction on DiNAT, encoder of our network, following the work presented in \cite{15}.

To begin with, DiNAT employs two 3 × 3 convolutional layers with a stride of 2 as a tokenizer to obtain a feature map with a resolution of one-fourth of the input image. Additionally, DiNAT utilizes a single 3 × 3 convolutional layer with a stride of 2 for downsampling between 
hierarchical levels, reducing the spatial resolution by half while doubling the number of channels.The resulting feature maps are thus of sizes $\frac{h}{4}\times\frac{w}{4}\times{c}$, $\frac{h}{8}\times \frac{w}{8}\times{2c}$, $\frac{h}{16}\times\frac{w}{16}\times{4c}$ and $\frac{h}{32}\times\frac{w}{32}\times{8c}$.

DiNAT adopts a straightforward stacking of DiNA layers, following a similar structural pattern as other commonly used Transformers. 
For simplicity, we keep notations limited to single dimensional NA and DiNA. 
Given input $X \in \mathbb{R}^{n \times d}$, whose rows are d-dimensional token vectors, and query and key linear projections of $X$, $Q$ and $K$, and relative positional biases between any two tokens $i$ and $j$, $B(i, j)$, $\delta$-dilated neighborhood attention weights for the $i$-th token with neighborhood size $k$, $\mathbf{A}_i^{(k, \delta)}$, is defined as the matrix multiplication of the $i$-th token’s query projection, and its k nearest neighboring tokens’ key projections with dilation value $\delta$: 

\begin{equation}
\mathbf{A}_i^{(k, \delta)}=\left[\begin{array}{c}
Q_i K_{\rho_1^\delta(i)}^T+B_{\left(i, \rho_1^\delta(i)\right)} \\
Q_i K_{\rho_2^\delta(i)}^T+B_{\left(i, \rho_2^\delta(i)\right)} \\
\vdots \\
Q_i K_{\rho_k^\delta(i)}^T+B_{\left(i, \rho_k^\delta(i)\right)}
\end{array}\right],
\end{equation}
where $\rho_j^\delta(i)$ denotes $i$'s $j$-th nearest neighbor with dilation value $\delta$. The $\delta$-dilated neighboring values for the $i$-th token is similarly defined with neighborhood size $k$, $\mathbf{V}_i^{(k, \delta)}$:
\begin{equation}
\mathbf{V}_i^{(k, \delta)}=\left[\begin{array}{llll}
V_{\rho_1^\delta(i)}^T & V_{\rho_2^\delta(i)}^T & \cdots & V_{\rho_k^\delta(i)}^T
\end{array}\right]^T,
\end{equation}
where $V$ is a linear projection of $X$. DiNA output for the $i$-th token neighborhood size $k$ with dilation value $\delta$ is then defined as:
\begin{equation}
\operatorname{DiNA}_k^\delta(i)=\operatorname{softmax}\left(\frac{\mathbf{A}_i^{(k, \delta)}}{\sqrt{d_k}}\right) \mathbf{V}_i^{(k, \delta)},
\end{equation}
where $\sqrt{d}$ is the scaling parameter, and $d$ is the embedding dimension. This operation is repeated for every pixel in the feature map.

Summary of DiNAT configuartions and dilation values will be provided in the supplementary material.

\subsection{SCAF-MLA Decoder}

Decoders play a critical role in various vision tasks. Taking inspiration from multilevel feature fusion techniques employed in vision tasks, we propose a novel decoder, SCAF-MLA, to effectively utilize numerous channels in the feature maps output from transformer-based encoder. SCAF-MLA enables the supervision on multiple levels, and learns rich hierarchical features, thus enhances the performance of edge detection. Besides, SCAF-MLA Decoder is more computationally efficient, without the commonly employed PPM\cite{36} and bottom-up path\cite{23,11}, while experimental results demonstrate that our designed decoder achieves more superior performance.

\noindent
\textbf{SCAFM.}

\begin{figure}[t]
    \centering
    \includegraphics[width=1\linewidth]{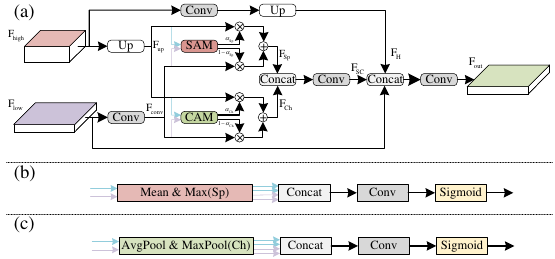}
    \caption{(a) The detailed architecture of the SCAFM. (b) The Spatial Attention Module(SAM). (c) The Channel Attention Module(CAM).}
    \label{fig:fig3}
\end{figure}

Inspired by UAFM\cite{34} in multi-level features fusing, we propose the Spatial and Channel Attention Fusion Module (SCAFM) as the main component of the SCAF-MLA. SCAFM is designed to extract both spatial and channel features, concurrently preserving the distinctive attributes of the current level while capturing higher-level features. The architecture of SCAFM is depicted in Figure \ref{fig:fig3}. SCAFM consists of a spatial attention module (SAM) and a channel attention module (CAM) to compute inter-spatial and inter-channel weights, denoted as $\alpha_{\text{Sp}}$ and $\alpha_{\text{Ch}}$, respectively. Specifically,  the upper-level feature is denoted as ${F}_{high}$ and the current-level feature as ${F}_{low}$. To begin with, bilinear interpolation is employed to upsample ${F}_{high}$ to the same size as ${F}_{low}$. Subsequently, convolutional operations is utilized to increase the channels of ${F}_{low}$ to match those of ${F}_{high}$, denoted as ${F}_{conv}$. For $\alpha_{\text{Sp}}$, we start by performing mean and max operations along the channel dimension on ${F}_{up}$ and ${F}_{conv}$, resulting in the generation of four features, each with a dimension of $\mathbb{R}^{1 \times H \times W}$. Subsequently, these four features are concatenated and processed through convolutional and sigmoid operations, yielding $\alpha_{\text{Sp}} \in \mathbb{R}^{1 \times H \times W}$. This process can be represented by thefollowing equations:
\begin{align}
\begin{split}
    F_{up}&=Up(F_{high}), \\
    F_{conv}&=Conv(F_{low}), \\
    \alpha_{\text{Sp}}&=Sigmoid(Conv(Cat(Mean(F_{up}), \\
    &Max(F_{up}),Mean(F_{conv}), \\
    &Max(F_{conv})))),
\end{split}
\end{align}

Regarding $\alpha_{\text{Ch}}$, average pooling and max pooling operations are applied on ${F}_{up}$ and ${F}_{conv}$, generating four features with dimensions $\mathbb{R}^{C \times 1 \times 1}$. Then, these features are concatenated and subjected to convolutional and sigmoid operations, generating $\alpha_{\text{Ch}} \in \mathbb{R}^{C \times 1 \times 1}$ , described as:
\begin{align}
\begin{split}
    \alpha_{\text{Ch}}&=Sigmoid(Conv(Cat(AvgPool(F_{up}), \\
    &MaxPool(F_{up}),AvgPool(F_{conv}), \\
    &MaxPool(F_{conv})))),
\end{split}
\end{align}
The input features are then fused with the generated weights $\alpha_{\text{Sp}}$ and $\alpha_{\text{Ch}}$ through multiplication and addition operations, resulting in features $F_{Sp}$ and $F_{Ch}$. Subsequently, these features are concatenated and convolved, generating $F_{SC} \in \mathbb{R}^{C \times H \times W}$. Then, we perform convolutions on $F_{high}$ followed by upsampling, generating features $F_H$ with dimensions $\mathbb{R}^{C \times H \times W}$. Afterwards, $F_H$, $F_{SC}$, and $F_{low}$ are concatenated and convolved to obtain the fused feature $F_{out} \in \mathbb{R}^{C \times H \times W}$. The aforementioned process can be described as:
\begin{align}
\begin{split}
    F_{up}=&Up(Conv(F_{high})), \\
    F_{Sp}=&F_{up}\cdot\alpha_{\text{Sp}}+F_{conv}\cdot(1-\alpha_{\text{Sp}}), \\
    F_{Ch}=&F_{up}\cdot\alpha_{\text{Ch}}+F_{conv}\cdot(1-\alpha_{\text{Ch}}), \\
    F_{SC}=&Conv(Cat(F_{Sp},F_{Ch})), \\
    F_{out}=&Conv(Cat(F_{SC},F_H,F_{low})), \\
\end{split}
\end{align}

\noindent
\textbf{Pre-fusion.} Most previous detectors\cite{10,24} fuse feature maps from different layers only after reducing their channels to 1, resulting in insufficient feature integration. Inspried by EDTER\cite{11}, which fuses the feature maps with a larger number of channels, we apply one, two, and three 3×3 convolutions to the feature maps $X_1$, $X_2$, and $X_3$ outputted by SCAFM respectively, reducing their channels to match that of $X_0$, rather than reducing to 1. We then use bilinear interpolation to upsample $X_1$, $X_2$, and $X_3$ to match $X_0$. Subsequently, we perform a concatenation operation on these four feature maps, and further reduce the channels to 1 using two 3×3 convolutions and one 1×1 convolution. Finally, we upsample the 1-channel feature map using bilinear interpolation and compute the sigmoid function to obtain the edge map $E \in \mathbb{R}^{1 \times H \times W}$.

\subsection{Loss Function}

We employ the loss function proposed in \cite{9} for the 4 side edge maps and 1 primary edge map. Given an edge map $E$ and its corresponding ground truth $Y$, the loss function is computed as follows:
\begin{align}
\begin{split}
\ell(E, Y)=&-\sum_{i, j}(Y_{i, j} \alpha \log \left(E_{i, j}\right) \\
+&\left(1-Y_{i, j}\right)(1-\alpha) \log \left(1-E_{i, j}\right)),
\end{split}
\end{align}
where $E_{i,j}$ and $Y_{i,j}$ are the $(i,j)^{th}$ element of matrix $E$ and $Y$, respectively. $\alpha = \frac{|Y^{-}|}{|Y^{-}| + |Y^{+}|}$ represents the percentage of negative pixel samples, with $|Y^{+}|$ and $|Y^{-}|$ denoting the number of positive and negative sample pixels, respectively. Since BSDS500 dataset is annotated by multiple annotators, we first normalize the multiple annotations into edge probability maps within the range of [0, 1]. Then, if the probability of a pixel is greater than a threshold value $\eta$, it is labeled as a positive sample; otherwise, it is labeled as a negative sample.

After the concatenation operation of the four feature maps output from our decoder, two 3×3 convolutions and one 1×1 convolution are applied to reduce the dimension of the concatenated feature maps. Similarly, a 3×3 convolution and a 1×1 convolution are applied to reduce the dimension of the four side feature maps. Subsequently, a sigmoid operation is performed on each of these reduced-dimensional feature maps to generate primary edge map and four side edge maps, denoted as $\mathcal{E}$, $\mathcal{S}_1$, $\mathcal{S}_2$, $\mathcal{S}_3$, and $\mathcal{S}_4$. We calculate the loss for both primary edge map and side edge maps to introduce additional supervision. To sum up, the overall loss function is as follows:
\begin{equation}
\mathcal{L}=\mathcal{L}^{\mathcal{E}}+\lambda \mathcal{L}^\mathcal{S}=\ell(\mathcal{E}, Y)+\lambda \sum_{k=1}^4 \ell\left(\mathcal{S}^k, Y\right),
\end{equation}
$\mathcal{L}^\mathcal{E}$ and $\mathcal{L}^\mathcal{S}$ represent the losses for the primary edge map $\mathcal{E}$ and the side edge maps $\mathcal{S}_1, \mathcal{S}_2, \mathcal{S}_3, \mathcal{S}_4$, respectively. Meanwhile, $\lambda$ denotes the weight that balances $\mathcal{L}^\mathcal{E}$ and $\mathcal{L}^\mathcal{S}$. Based on \cite{11} and our experimental observations, we set $\lambda$ to 0.4.

\section{Experiments}

\subsection{Datasets}

Two mainstream datasets are used to evaluate our proposed EdgeNAT, namely, BSDS500 and NYUDv2.

\noindent
\textbf{BSDS500}\cite{1} consists of 500 RGB images, with 200 for training, 100 for validation, and 200 for testing. Similar to \cite{9,10}, the dataset is augmented to 28,800 images by flipping, scaling, and rotating. PASCAL VOC Context dataset\cite{38} is used as additional training data and its 10,103 training images are also augmented to 20,206 by flipping, as in most previous works\cite{10,23}. The model is pre-trained with the augmented PACSAL VOC Context dataset and then fine-tuned with the 300 training and validation images of BSDS500 dataset, and is evaluated on 200 testing images.

\noindent
\textbf{NYUDv2}\cite{35} consists of 1449 labeled pairs of aligned RGB and depth images, with 381 training images, 414 validation images, and 654 testing images. As in \cite{9,10},the training and validation sets are combined and augmented to train the model.

\subsection{Implementation Details}
Our EdgeNAT is implemented with PyTorch and is based on mmsegmentation\cite{37} and NATTEN\cite{14}. 
We use the pre-trained weights of DiNAT\cite{15} to initialize EdgeNAT’s transformer blocks. 
To generate binary edge maps, for BSDS500, we set the threshold $\eta$ to 0.3 to select positive samples. 
For NYUDv2, there is only one annotation per picture, so there is no need to set the threshold $\eta$.

We use the AdamW optimizer and train for 40k iterations using a cosine decay learning rate scheduler, where the first 15k iterations warm up the learning rate in a linear manner, and the remaining ones are decayed according to the scheduler. The initial learning rate is 0 and a preset learning rate is set to 6e-5. For BSDS500, we set its batch size to 8, and for NYUDv2, we set its batch size to 4.

All experiments were conducted on RTX 4090 GPU. The training of the L model of EdgeNAT (472.38MB) takes 6 hours, far more efficient than Transformer-based model EDTER (468.84MB)\cite{11}, which takes 26.4 hours. The inference runs at 20.87 FPS on RTX 4090, nearly ten times the speed of EDTER on V100 (2.2 FPS). During training, since our model is a one-stage edge detection model, for 320×320 images, the GPU memory requirement is about 20GB, 2/3 of EDTER(29GB).

Optimal Dataset Scale (ODS) and Optimal Image Scale (OIS) are two metrics for all datasets. Before evaluation, we perform non-maximum suppression on the predicted edge maps. For the maximum allowed tolerance distance between the detected edge and ground truth, we set it to 0.0075 for BSDS500 and to 0.011 for NYUDv2 as in previous works.
\subsection{Ablation Study}
Ablation experiments are performed on the BSDS500 data set to verify the effectiveness of our proposed decoder. Specifically, we first compare the effect of pre-fusion (reduce the channels of feature map to C) and final-fusion (reduce the channels of feature map to 1); then the effect of bottom-up path is also verified. From the quantitative results shown in Table \ref{tab:effectiveness_of_the_pre-fusion_and_bottom-up_path}, it is clear that regardless of pre-fusion or final-fusion, Bottom-up Path has negative effects on edge detection performance, indicating it is not suitable for DiNAT. For edge detection models with relatively large number of feature map channels, pre-fusion without PPM will be a better choice.
\begin{table}[ht]
    \centering
    \begin{tabular}{c|cc}
        \toprule
        ODS / OIS  & Final-fusion & Pre-fusion \\
        \hline
        Bottom-up Path  & 0.838 / 0.852          & 0.839 / 0.856        \\
        -               & 0.838 / 0.853          & 0.840 / 0.856        \\
        \bottomrule
    \end{tabular}
    \caption{Ablation study of the effectiveness of the pre-fusion and bottom-up path on BSDS500. All results are computed with a single-scale input without additional training data.}
    \label{tab:effectiveness_of_the_pre-fusion_and_bottom-up_path}
\end{table}
\begin{table}[ht]
    \centering
    \begin{tabular}{cc|cc}
        \toprule
        PPM        &   SCAFM      &   ODS &   OIS     \\
        \midrule
        \ding{55}  & \ding{55}    & 0.837 &   0.853   \\
        \ding{51}  & \ding{55}    & 0.836 &   0.851   \\
        \ding{55}  & \ding{51}    & 0.843 &   0.859   \\
        \ding{51}  & \ding{51}    & 0.841 &   0.856   \\
        \bottomrule
    \end{tabular}
    \caption{Ablation study on the effectiveness of PPM and SCAFM. All results are computed with a single-scale input without additional training data.}
    \label{tab:2}
\end{table}

Next, the effectiveness of PPM\cite{36} and our proposed SCAFM are verified and compared. The quantitative results shown in Table \ref{tab:2} demonstrates that SCAFM works best without PPM, achieving best ODS and OIS score, 84.3\% and 85.9\% respectively. In summary, we will use the SCAF-MLA decoder without Bottom-up Path and PPM for the next experiments.


\subsection{Network Scalability}
\begin{figure}[t]
    \centering
    \includegraphics[width=\linewidth]{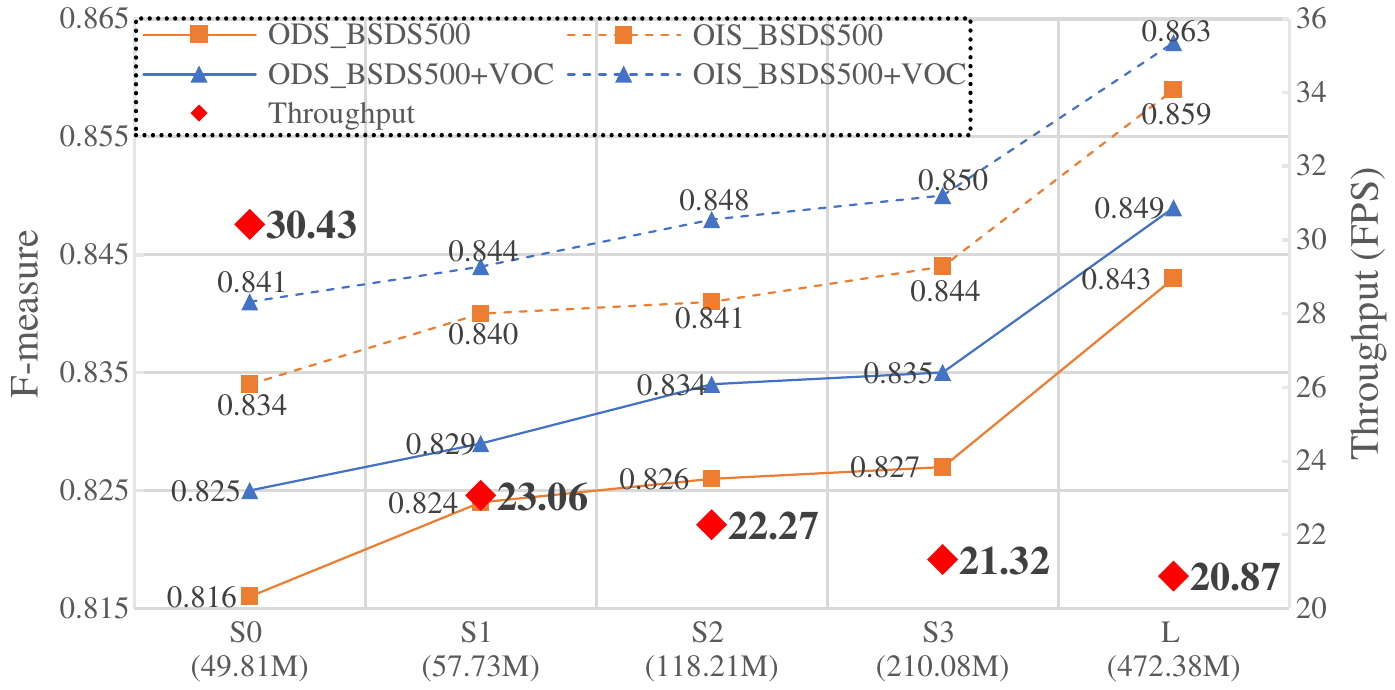}
    \caption{Exploration on the scalability of EdgeNAT. Bottom row shows the number of parameters for each model. The models are trained with or without PASCAL VOC dataset. All results are computed with a single-scale input.}
    \label{fig:NS}
\end{figure}
EdgeNAT-L has a relatively large amount of parameters (472.38MB). In order to adapt to different application scenarios, we conduct scalability experiments on different model sizes. The configuration settings of the encoder of the L, S0, S1, S2, and S3 variants of our EdgeNAT are the same as those of the Large, Mini, Tiny, Small, and Base versions of the DiNAT\cite{15}. Extensive experiments are conducted to study the scalability and throughput of EdgeNAT variants. The result is shown in Figure \ref{fig:NS}. The models are all trained using the BSDS500 training and validation sets with or without PASCAL VOC, and evaluated with the BSDS500 test set. As expected, when the size of our model decreases, the ODS and OIS will decrease accordingly, and the throughput increases.

It is worth noting that the processing speed of the S0 model is much higher than that of other models. This should be contributed to the fact that its third level has only 6 layers, while the others models have 18 layers. The ODS and OIS of the L model are much higher than other models, due to the fact that the encoder is pre-trained on ImageNet-22K, while encoder of other models are pre-trained on ImageNet-1K. The results of multi-scale input experiment of S0, S1, S2, and S3 models as well as their visualization results will be provided in the supplementary material. 

\begin{figure*}[ht]
  \centering
  \includegraphics[width=\textwidth]{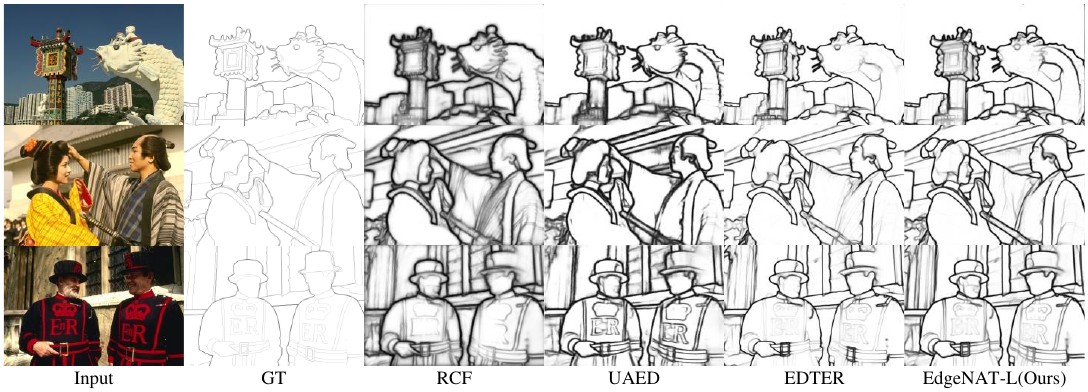}
  \caption{Qualitative comparisons on three challenging samples in the testing set of BSDS500. It is interesting to notice that in the third example, the edge of the hat on the right is completed by our L model, though the hat edge is hard to distinguish even for human eyes. This unprecedented phenomenon demonstrates our model has better global semantic understanding than previous works.}
  \label{fig:Qualitative_comparisons_BSDS500}
\end{figure*}

\subsection{Comparison with State-of-the-arts}
\begin{figure}[ht]
  \centering
  \includegraphics[width=\linewidth]{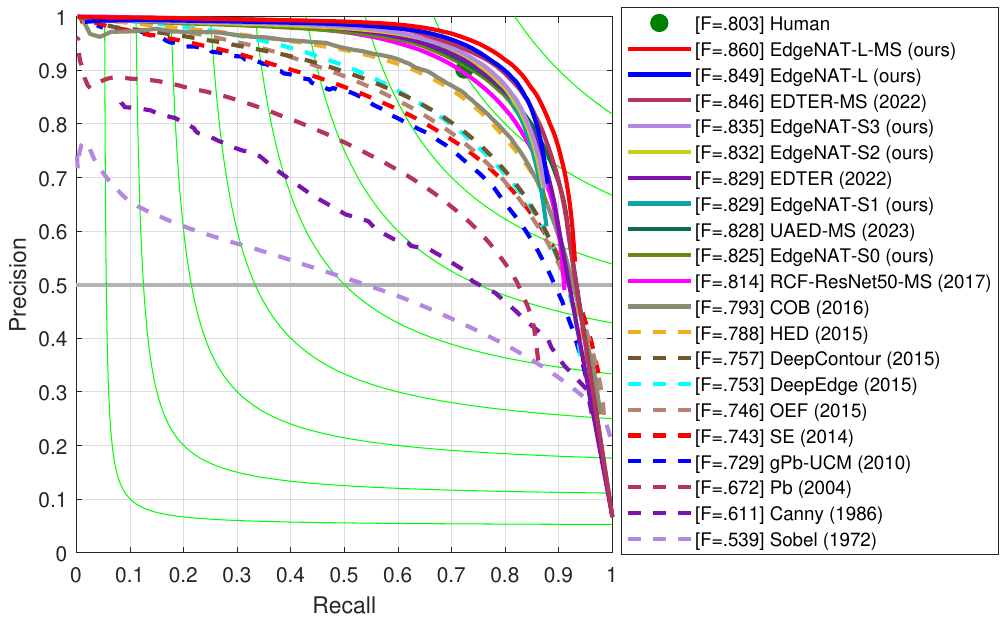}
  \caption{The precision-recall curves on BSDS500.}
  \label{fig:pr_bsds}
\end{figure}

\noindent
\textbf{On BSDS500 dataset.} We compare our L model with {\it traditional detectors} such as Canny\cite{6}, gPb-UCM\cite{39}, SCG\cite{40}, SE\cite{41} and  OEF\cite{42}, and {\it CNN-based detector} such as DeepEdge\cite{7}, DeepContour\cite{43}, HED\cite{9}, Deep Boundary\cite{44}, CEDN\cite{45}, RDS\cite{46}, AMH-Net\cite{26}, RCF\cite{10}, CED\cite{21}, LPCB\cite{20}, BDCN\cite{23}, DSCD\cite{22}, PiDiNet\cite{24}, UAED\cite{28} and PEdger\cite{27}, and {\it transformer-based detector} such as EDTER\cite{11}. The results are summarized in Table \ref{tab:3} and Figure \ref{fig:pr_bsds}, respectively. We notice that our L model, trained on the BSDS500 dataset, achieves an ODS of 84.3\% with single-scale inputs, outperforming all competing detectors. Furthermore, when employing multi-scale inputs, our method achieves an even higher ODS of 85.5\%. By utilizing additional training data and adopting multi-scale input (following the configurations of RCF, EDTER, {\it etc.}), our method attains 86.0\%(ODS), 87.6\%(OIS), which clearly demonstrate the superiority of our method over all existing state-of-the-art edge detectors. Several qualitative results are presented in Figure \ref{fig:Qualitative_comparisons_BSDS500}. It can be observed that our proposed EdgeNAT demonstrates a distinct advantage in terms of prediction quality. The generated outputs exhibit clear and exact edge predictions, further validating the efficacy of our method.
\begin{table}[ht]
    \centering
    \begin{tabular}{c|l|c|cc}
        \toprule
        \multicolumn{2}{c|}{Method}      & Pub.'Year     & ODS      & OIS    \\
        \midrule
        \multirow{5}{*}{\rotatebox{90}{Traditional}} &Canny       & PAMI'86       & 0.611    & 0.676  \\
        &gPb-UCM        & PAMI'10       & 0.729    & 0.755  \\
        &SCG            & NeurIPS'12    & 0.739    & 0.758  \\
        &SE             & PAMI'14       & 0.743    & 0.764  \\
        &OEF            & CVPR'15       & 0.746    & 0.770  \\
        \midrule
        \multirow{17}{*}{\rotatebox{90}{CNN-based}} &DeepEdge     & CVPR'15       & 0.753    & 0.772  \\
        &DeepContour    & CVPR'15       & 0.757     & 0.776     \\
        &HED            & ICCV'15       & 0.788     & 0.808     \\
        &Deep Boundary\textdagger\textdaggerdbl  & ICLR'15       & 0.789     & 0.811     \\
        &CEDN           & CVPR'16       & 0.788     & 0.804     \\
        &RDS            & CVPR'16       & 0.792     & 0.810     \\
        &AMH-Net        & NeurIPS'17    & 0.798     & 0.829     \\
        &RCF\textdagger\textdaggerdbl            & CVPR'17       & 0.811     & 0.830     \\
        &CED\textdagger            & CVPR'17       & 0.815     & 0.833     \\
        &LPCB\textdagger\textdaggerdbl           & ECCV'18       & 0.815     & 0.834     \\
        &BDCN\textdagger\textdaggerdbl           & CVPR'19       & 0.828     & 0.844     \\
        &DSCD\textdagger\textdaggerdbl           & ACMMM'20      & 0.822     & 0.859     \\
        &PiDiNet\textdagger        & ICCV'21       & 0.807     & 0.823     \\
        &UAED\textdagger\textdaggerdbl           & CVPR'23       & 0.844     & 0.864     \\
        &PEdger-large\textdagger         & ACMMM'23      & 0.823     & 0.841     \\
        \midrule
        \multirow{8}{*}{\rotatebox{90}{Transformer-based}} &EDTER    & \multirow{4}{*}{CVPR'22}       & 0.824    & 0.841  \\
        &EDTER\textdagger    &        & 0.832    & 0.847  \\
        &EDTER\textdaggerdbl    &        & 0.840    & 0.858  \\
        &EDTER\textdagger\textdaggerdbl    &        & 0.848    & 0.865  \\
        \cline{2-5}
        &EdgeNAT-L   & \multirow{4}{*}{Ours}          & 0.843     & 0.859     \\
        &EdgeNAT-L\textdagger    &      & 0.849     & 0.863     \\
        &EdgeNAT-L\textdaggerdbl     &      & \textbf{\textcolor{blue}{0.855}}     & \textbf{\textcolor{blue}{0.870}}     \\
        &EdgeNAT-L\textdagger\textdaggerdbl &      & \textbf{\textcolor{red}{0.860}}     & \textbf{\textcolor{red}{0.876}}     \\
        \bottomrule
    \end{tabular}
    \caption{Results on BSDS500 testing set. The best two results are highlighted in \textbf{\textcolor{red}{red}} and \textbf{\textcolor{blue}{blue}}, respectively, and same for other tables. \textdagger \ means training with extra PASCAL VOC data, and \textdaggerdbl \ is the multi-scale testing.}
    \label{tab:3}
\end{table}

\noindent
\textbf{On NYUDv2 dataset.} We conduct experiments on three types of inputs (RGB, HHA, and RGB-HHA). The RGB-HHA results are obtained by averaging the edge detections from RGB and HHA. We compare our L model with deep learning-based detectors, including HED\cite{9}, COB\cite{25}, RCF\cite{10}, AMH-Net\cite{26}, LPCB\cite{20}, BDCN\cite{23}, PiDiNet\cite{24}, PEdger\cite{27}, and EDTER\cite{11}. All results are based on single-scale inputs. The results are shown in Table \ref{tab:4}. It can be observed that our L model achieveds ODS of 78.9\%, 72.6\%, and 79.4\% for RGB, HHA, and RGB-HHA, respectively, surpassing the second-best method by 1.5\%, 0.9\% and 1.0\%, respectively. Furthermore, our approach also attains the highest OIS among all the evaluated methods.
The results of our other models, and the precision-recall curves, will be presented in the supplementary material.

\begin{table}[ht]
    \centering
    \resizebox{\columnwidth}{!}{
    \begin{tabular}{l|cc|cc|cc}
        \toprule
        \multirow{2}{*}{Method}  & \multicolumn{2}{c}{RGB} & \multicolumn{2}{|c}{HHA} & \multicolumn{2}{|c}{RGB-HHA} \\
        \cline{2-7}
        & ODS & OIS & ODS & OIS & ODS & OIS \\
        \midrule
        HED     & 0.720 & 0.734 & 0.682 & 0.695 & 0.746 & 0.761 \\
        COB     & -     & -     & -     & -     & \textbf{\textcolor{blue}{0.784}} & \textbf{\textcolor{blue}{0.805}} \\
        RCF     & 0.729 & 0.742 & 0.705 & 0.715 & 0.757 & 0.771 \\
        AMH-Net & 0.744 & 0.758 & \textbf{\textcolor{blue}{0.717}} & \textbf{\textcolor{blue}{0.729}} & 0.771 & 0.786 \\
        LPCB    & 0.739 & 0.754 & 0.707 & 0.719 & 0.762 & 0.778 \\
        BDCN    & 0.748 & 0.763 & 0.707 & 0.719 & 0.765 & 0.781 \\
        PiDiNet & 0.733 & 0.747 & 0.715 & 0.728 & 0.756 & 0.773 \\
        PEdger  & 0.742 & 0.757 & -     & -     & -     & -     \\
        \midrule
        EDTER   & \textbf{\textcolor{blue}{0.774}} & \textbf{\textcolor{blue}{0.789}} & 0.703 & 0.718 & 0.780 & 0.797 \\
        EdgeNAT-L & \textbf{\textcolor{red}{0.789}} & \textbf{\textcolor{red}{0.803}} & \textbf{\textcolor{red}{0.726}} & \textbf{\textcolor{red}{0.741}} & \textbf{\textcolor{red}{0.794}} & \textbf{\textcolor{red}{0.808}} \\
        \bottomrule
    \end{tabular}}
    \caption{Quantitative comparisons on NYUDv2. All results are computed with a single scale input.}
    \label{tab:4}
\end{table}

\section{Conclusion}
Our contributions are summarized as follows: firstly, we introduce DiNAT as the encoder, which enables our proposed edge detector not only more accurate than current SOTA EDTER, but also ten times faster than it. Secondly, we propose SCAFM, a module that concatenates spatial attention and channel attention, to generate richer and more accurate feature representation for the decoder. Thirdly, we design five version of models with different parameter sizes to adapt to complex and diverse application scenarios and conduct extensive experiments on the BSDS500 and NYUDv2 datasets, demonstrating that EdgeNAT achieves superiority in both efficiency and accuracy.

\bibliography{EdgeNAT}
\cleardoublepage
\appendixpage
\appendix

In this appendices, we provide additional detailed information, including EdgeNAT configurations and dilation values, as well as more experimental results and their visualization on BSDS500\cite{1} and NYUDv2\cite{35}.

\section{EdgeNAT Configurations and Dilation Values}

Our EdgeNAT model, including the S0, S1, S2, S3 and L variants, utilizes a hierarchical backbone network architecture divided into four levels. Each level consists of a Downsampler (or Tokenizer) and multiple DiNAT blocks. The configurations and dilation values of these levels closely resemble those of the Mini, Tiny, Small, Base, and Large variants of DiNAT, as presented in Table \ref{tab:DiNAT configurations and dilation values.}.

\section{More Results on BSDS500}

\begin{figure}[ht]
    \centering
    \includegraphics[width=1\linewidth]{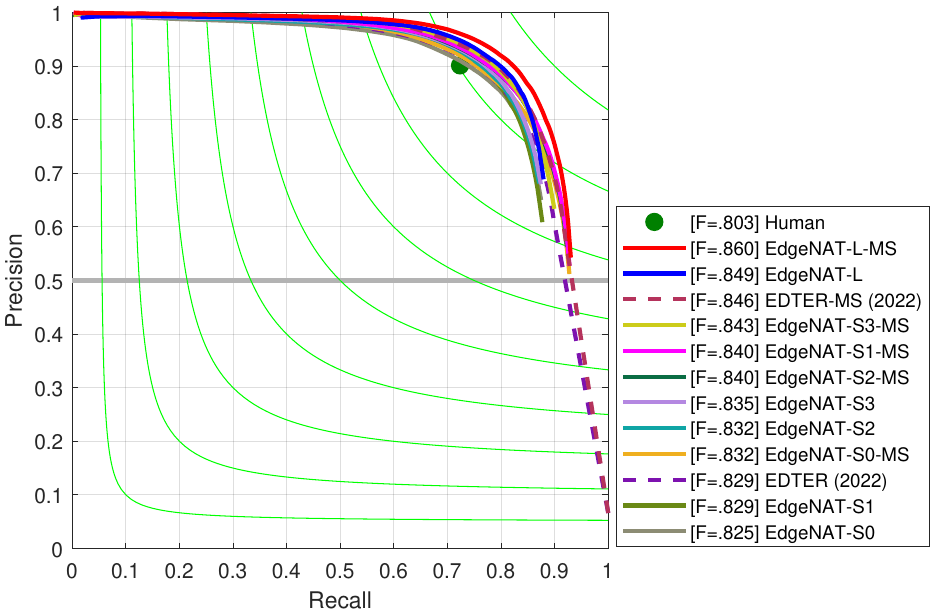}
    \caption{The Precision-Recall curves on BSDS500 for transformer-based detector}
    \label{fig:pr_bsds_variants}
\end{figure}

On BSDS500\cite{1} dataset, since the main text presents the experimental results with single-scale input, in the supplementary material, we conduct experiments on the S0, S1, S2, and S3 models with multi-scale inputs. 
Besides, the single-scale and multi-scale experiments with additional training data are also conducted. We compare our models with {\it traditional detectors} such as Canny\cite{6}, gPb-UCM\cite{39}, SCG\cite{40}, SE\cite{41} and  OEF\cite{42}, {\it CNN-based detector} such as DeepEdge\cite{7}, DeepContour\cite{43}, HED\cite{9}, Deep Boundary\cite{44}, CEDN\cite{45}, RDS\cite{46}, AMH-Net\cite{26}, RCF\cite{10}, CED\cite{21}, LPCB\cite{20}, BDCN\cite{23}, DSCD\cite{22}, PiDiNet\cite{24}, UAED\cite{28} and PEdger\cite{27}, and {\it transformer-based detector} such as EDTER\cite{11}. 
The results are presented in Table \ref{tab:compare_BSDS500_tab}. Moreover, Figure \ref{fig:pr_bsds_variants} shows the Precision-Recall curves of transformer-based detector, Figure \ref{fig:Qualitative_results_of_EdgeNAT-L_on_BSDS500} provides qualitative results of the L model, Figure \ref{fig:Qualitative_comparisons_of_other_variants_on_BSDS500} shows the visual results for different variants of EdgeNAT, and Figure \ref{fig:Qualitative_comparisons_on_BSDS500} presents the visual results compared with other approaches. 
It is evident that even the variants of EdgeNAT with smaller parameter sizes yield highly competitive results.

\section{More Results on NYUDv2}

\begin{figure}
    \centering
    \includegraphics[width=\linewidth]{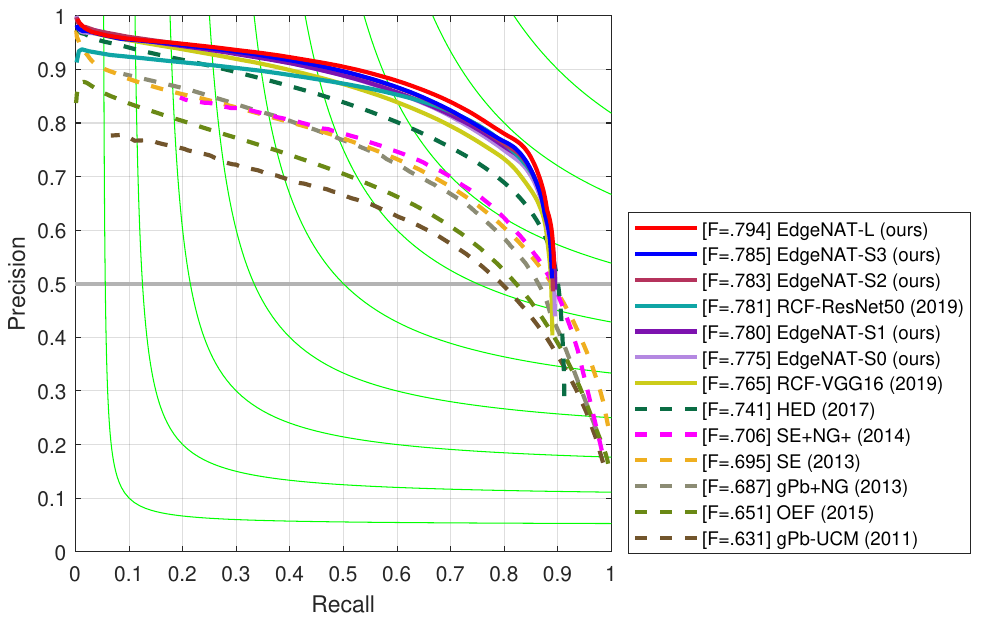}
    \caption{ The Precision-Recall curves on NYUDv2.}
    \label{fig:pr_nyud}
\end{figure}

\begin{figure*}[t]
  \centering
  \includegraphics[width=0.9\textwidth]{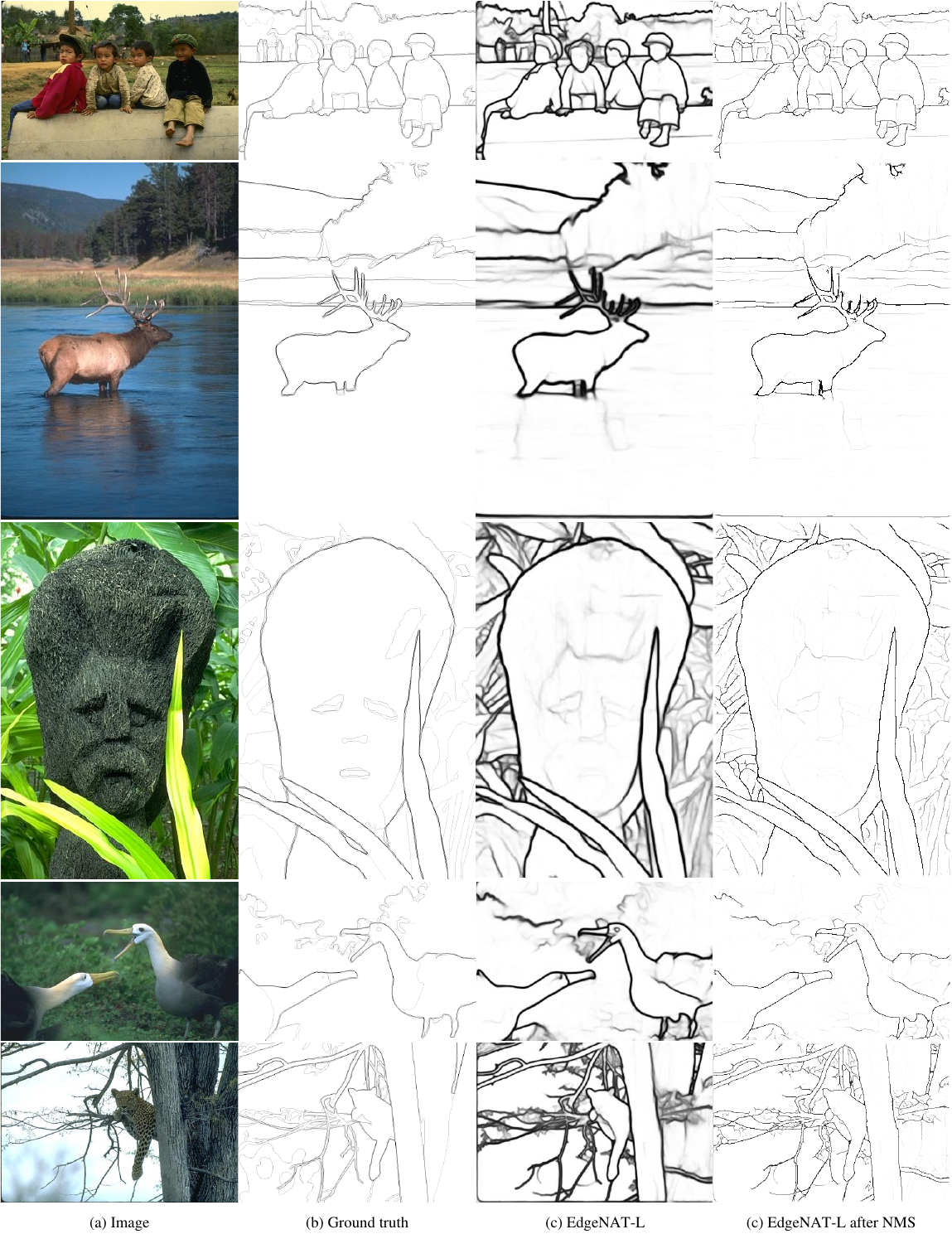}
  \caption{Qualitative results of EdgeNAT-L on BSDS500.}
  \label{fig:Qualitative_results_of_EdgeNAT-L_on_BSDS500}
\end{figure*}

\begin{figure*}[t]
  \centering
  \includegraphics[width=0.9\textwidth]{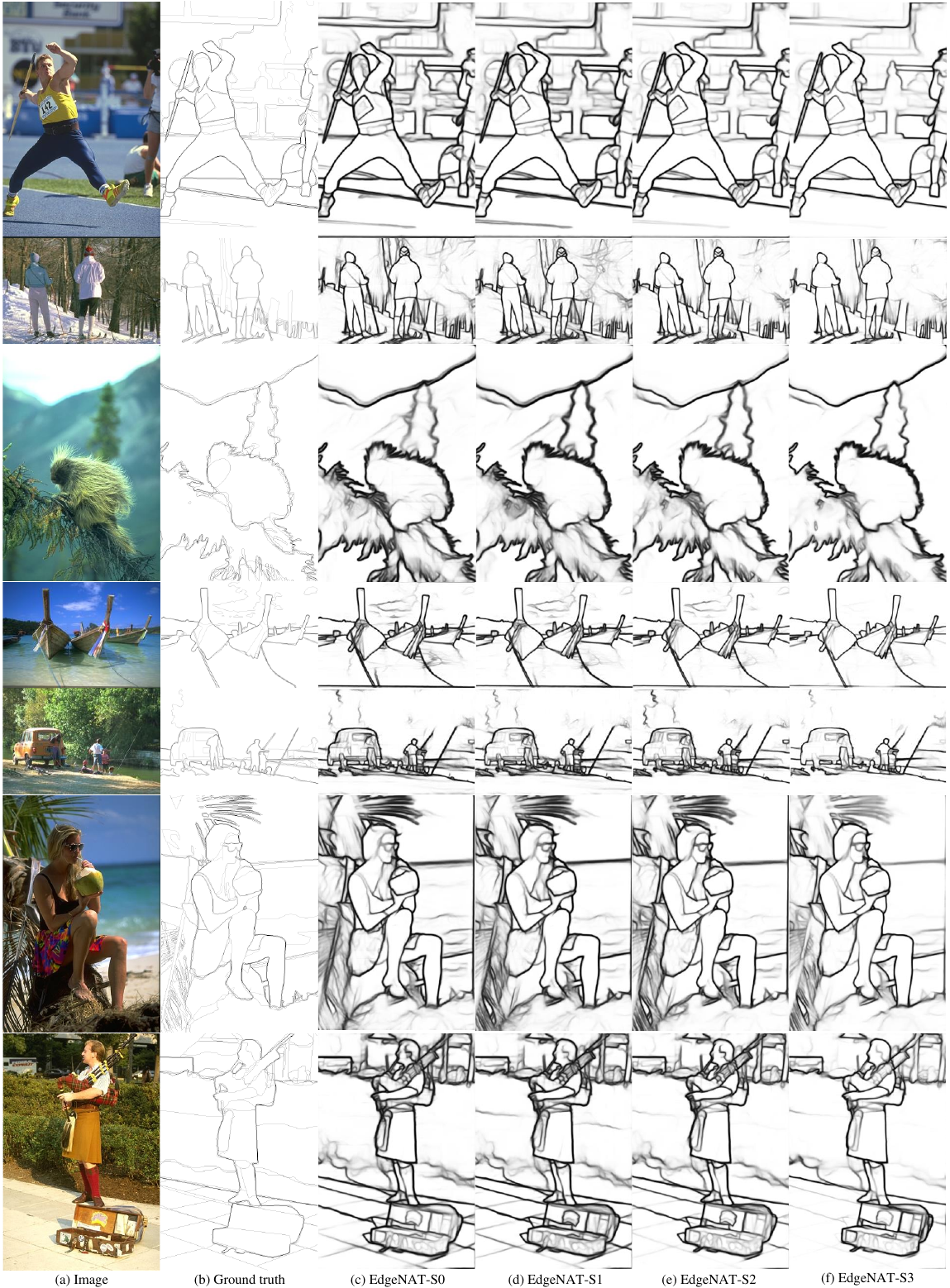}
  \caption{Qualitative comparisons of smaller variants on BSDS500.}
  \label{fig:Qualitative_comparisons_of_other_variants_on_BSDS500}
\end{figure*}

\begin{figure*}[t]
  \centering
  \includegraphics[width=0.9\textwidth]{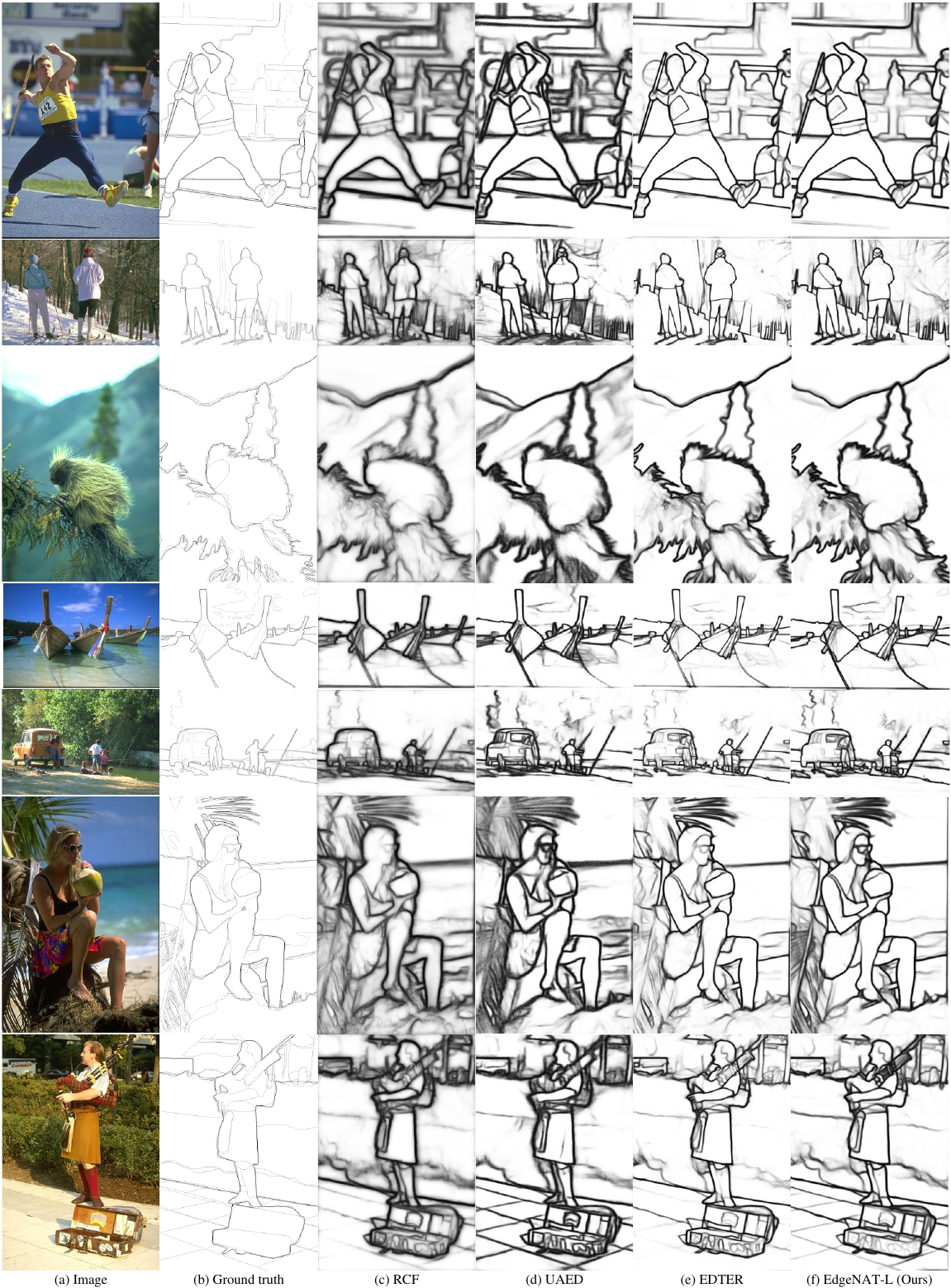}
  \caption{Qualitative comparisons with RCF, UAED and EDTER on BSDS500.}
  \label{fig:Qualitative_comparisons_on_BSDS500}
\end{figure*}

\begin{figure*}[t]
  \centering
  \includegraphics[width=0.9\textwidth]{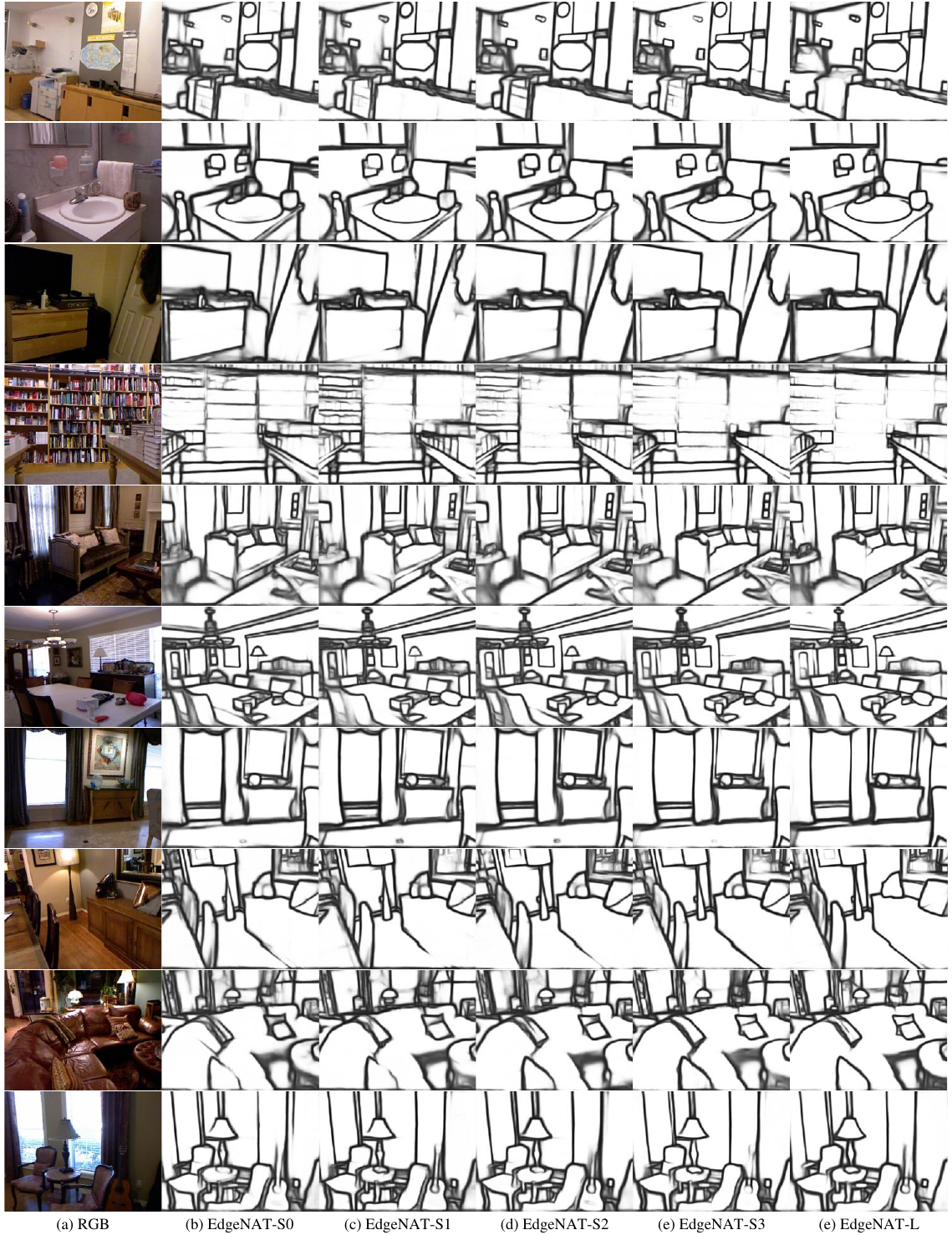}
  \caption{Qualitative comparisons of all variants on NYUDv2 with RGB inputs.}
  \label{fig:Qualitative_results_on_NYUDv2_RGB}
\end{figure*}

\begin{figure*}[t]
  \centering
  \includegraphics[width=0.9\textwidth]{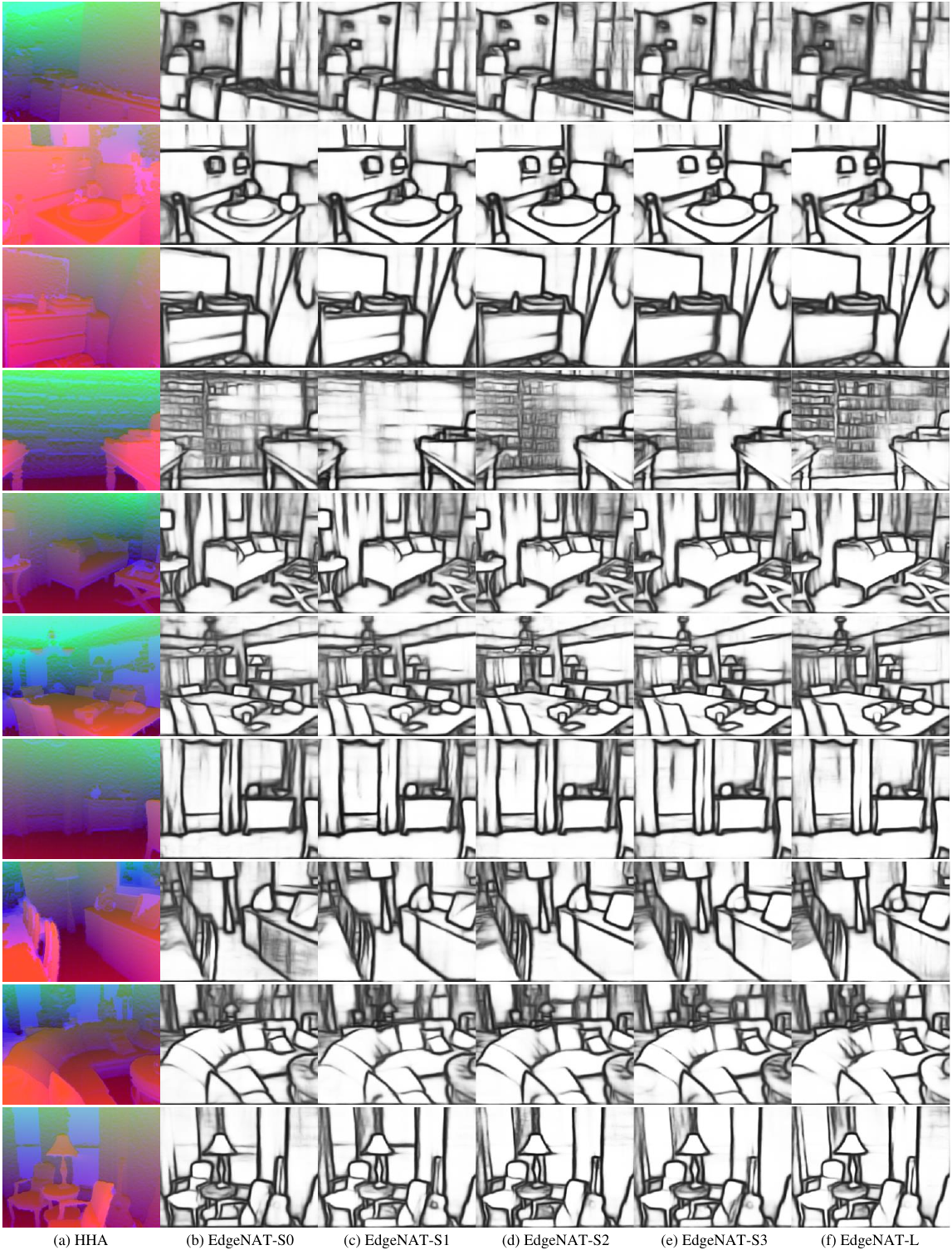}
  \caption{Qualitative comparisons of all variants on NYUDv2 with HHA inputs.}
  \label{fig:Qualitative_results_on_NYUDv2_HHA}
\end{figure*}

\begin{figure*}[t]
  \centering
  \includegraphics[width=0.9\textwidth]{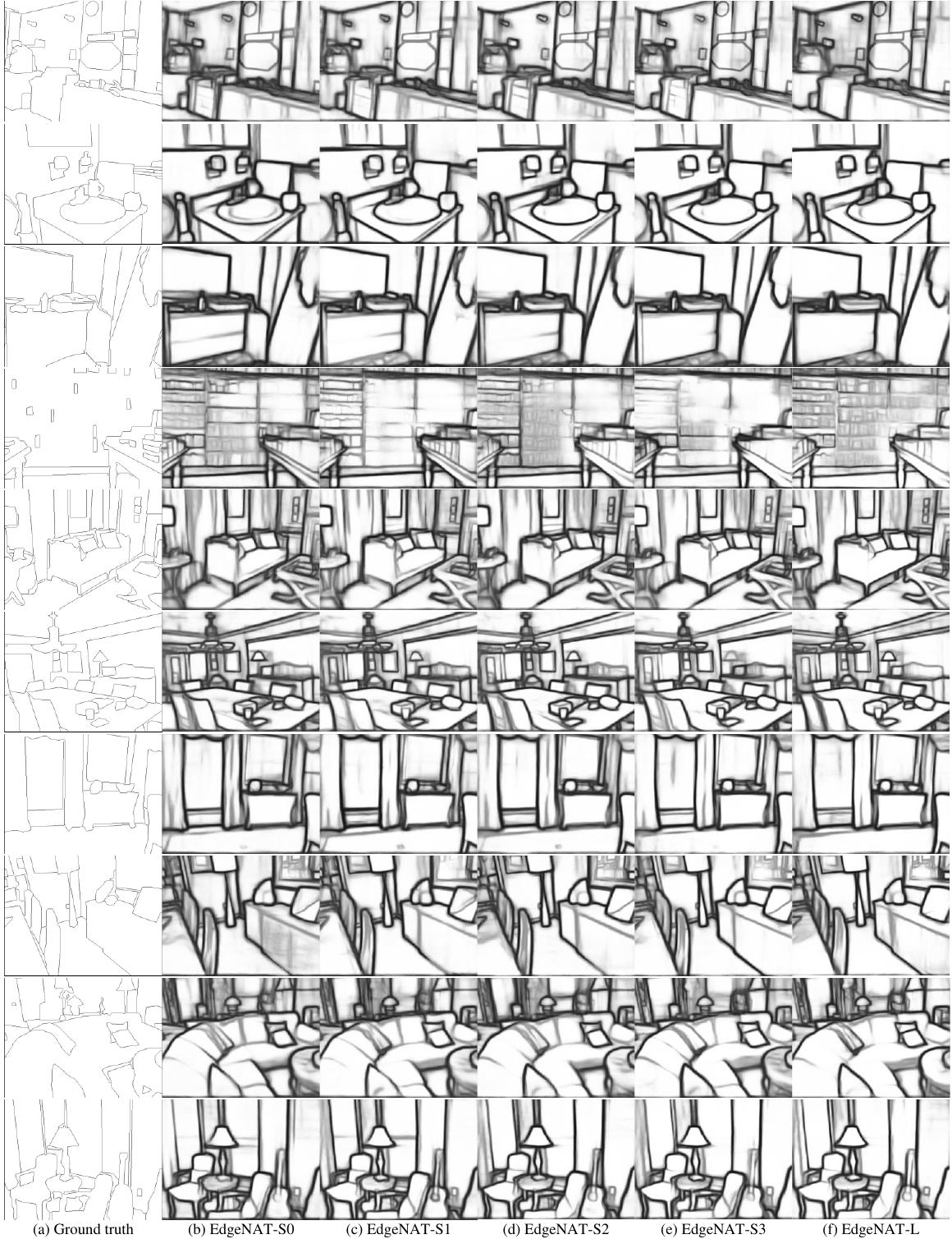}
  \caption{Qualitative comparisons of all variants on NYUDv2 with RGB+HHA inputs.}
  \label{fig:Qualitative_results_on_NYUDv2_RGB+HHA}
\end{figure*}

On the NYUDv2\cite{35} dataset, we conduct experiments on all variants of the model, exploring three types of inputs: RGB, HHA, and RGB-HHA. We compare our models with {\it traditional detectors} including gPb-UCM\cite{39}, gPb+NG\cite{49}, SE\cite{41}, SE+NG+\cite{50}, OEF\cite{42}, SemiContour\cite{51}, {\it CNN-based detectors} including HED\cite{9}, COB\cite{25}, RCF\cite{10}, AMH-Net\cite{26}, LPCB\cite{20}, BDCN\cite{23}, PiDiNet\cite{24}, PEdger\cite{27}, and {\it transformer-based detector} including EDTER\cite{11}. 
The results are shown in Table \ref{tab:Tab_quantitative_comparisons_on_NYUDv2}. The Precision-Recall curves of the experiment of RGB-HHA inputs are presented in Figure \ref{fig:pr_nyud}. The visualizations of RGB, HHA, and RGB+HHA results are presented in Figure \ref{fig:Qualitative_results_on_NYUDv2_RGB}, Figure \ref{fig:Qualitative_results_on_NYUDv2_HHA}, and Figure \ref{fig:Qualitative_results_on_NYUDv2_RGB+HHA}, respectively.

\vfill
\begin{table*}[]
    \centering
    \resizebox{\textwidth}{!}{
    \begin{tabular}{cccccccc}
        \toprule
        Variant  & Layers per level  & Level 1 & Level 2 & Level 3 & Level 4 & Dim $\times$ Heads & MLP ratio\\
        \midrule
        S0  & 3, 4, 6, 5    & 1, 16, 1      & 1, 4, 1, 8      & 1, 2, 1, 3, 1, 4                                        & 1, 2, 1, 2, 1   & 32 $\times$ 2      & 3\\
        S1  & 3, 4, 18, 5   & 1, 16, 1      & 1, 4, 1, 8      & 1, 2, 1, 3, 1, 4, 1, 2, 1, 3, 1, 4, 1, 2, 1, 3, 1, 4    & 1, 2, 1, 2, 1   & 32 $\times$ 2      & 3\\
        S2  & 3, 4, 18, 5   & 1, 16, 1      & 1, 4, 1, 8      & 1, 2, 1, 3, 1, 4, 1, 2, 1, 3, 1, 4, 1, 2, 1, 3, 1, 4    & 1, 2, 1, 2, 1   & 32 $\times$ 3      & 2\\
        S3  & 3, 4, 18, 5   & 1, 16, 1      & 1, 4, 1, 8      & 1, 2, 1, 3, 1, 4, 1, 2, 1, 3, 1, 4, 1, 2, 1, 3, 1, 4    & 1, 2, 1, 2, 1   & 32 $\times$ 4      & 2 \\
        L   & 3, 4, 18, 5   & 1, 20, 1      & 1, 5, 1, 10     & 1, 2, 1, 3, 1, 4, 1, 2, 1, 3, 1, 4, 1, 2, 1, 3, 1, 4    & 1, 2, 1, 2, 1   & 32 $\times$ 6      & 2\\
        \bottomrule
    \end{tabular}
    }
    \caption{DiNAT configurations and dilation values. Channels (heads $\times$ dim) double after every level until the final one. Kernel size is 7×7 in all variants.}
    \label{tab:DiNAT configurations and dilation values.}
\end{table*}

\begin{table}
    \centering
    \begin{adjustbox}{width=0.9\linewidth}
    \begin{tabular}{c|l|c|cc}
        \toprule
        \multicolumn{2}{c|}{Method}      & Pub.'Year     & ODS      & OIS    \\
        \midrule
        \multirow{5}{*}{\rotatebox{90}{Traditional}} &Canny       & PAMI'86       & 0.611    & 0.676  \\
        &gPb-UCM        & PAMI'10       & 0.729    & 0.755  \\
        &SCG            & NeurIPS'12    & 0.739    & 0.758  \\
        &SE             & PAMI'14       & 0.743    & 0.764  \\
        &OEF            & CVPR'15       & 0.746    & 0.770  \\
        \midrule
        \multirow{15}{*}{\rotatebox{90}{CNN-based}} &DeepEdge     & CVPR'15       & 0.753    & 0.772  \\
        &DeepContour    & CVPR'15       & 0.757     & 0.776     \\
        &HED            & ICCV'15       & 0.788     & 0.808     \\
        &Deep Boundary\textdagger\textdaggerdbl  & ICLR'15       & 0.789     & 0.811     \\
        &CEDN           & CVPR'16       & 0.788     & 0.804     \\
        &RDS            & CVPR'16       & 0.792     & 0.810     \\
        &AMH-Net        & NeurIPS'17    & 0.798     & 0.829     \\
        &RCF\textdagger\textdaggerdbl            & CVPR'17       & 0.811     & 0.830     \\
        &CED\textdagger            & CVPR'17       & 0.815     & 0.833     \\
        &LPCB\textdagger\textdaggerdbl           & ECCV'18       & 0.815     & 0.834     \\
        &BDCN\textdagger\textdaggerdbl           & CVPR'19       & 0.828     & 0.844     \\
        &DSCD\textdagger\textdaggerdbl           & ACMMM'20      & 0.822     & 0.859     \\
        &PiDiNet\textdagger        & ICCV'21       & 0.807     & 0.823     \\
        &UAED\textdagger\textdaggerdbl           & CVPR'23       & 0.844     & 0.864     \\
        &PEdger-large\textdagger         & ACMMM'23      & 0.823     & 0.841     \\
        \midrule
        \multirow{24}{*}{\rotatebox{90}{Transformer-based}} &EDTER    & \multirow{4}{*}{CVPR'22}       & 0.824    & 0.841  \\
        &EDTER\textdagger    &        & 0.832    & 0.847  \\
        &EDTER\textdaggerdbl    &        & 0.840    & 0.858  \\
        &EDTER\textdagger\textdaggerdbl    &        & 0.848    & 0.865  \\
        \cline{2-5}
        &EdgeNAT-S0   & \multirow{20}{*}{Ours}          & 0.816     & 0.834     \\
        &EdgeNAT-S0\textdagger    &      & 0.825     & 0.841     \\
        &EdgeNAT-S0\textdaggerdbl     &      & 0.829     & 0.846     \\
        &EdgeNAT-S0\textdagger\textdaggerdbl &      & 0.832     & 0.850     \\
        \cline{2-2}
        \cline{4-5}
        &EdgeNAT-S1   &          & 0.824     & 0.840     \\
        &EdgeNAT-S1\textdagger    &      & 0.829     & 0.844     \\
        &EdgeNAT-S1\textdaggerdbl     &      & 0.831     & 0.848     \\
        &EdgeNAT-S1\textdagger\textdaggerdbl &      & 0.840     & 0.858     \\
        \cline{2-2}
        \cline{4-5}
        &EdgeNAT-S2   &           & 0.826     & 0.841     \\
        &EdgeNAT-S2\textdagger    &      & 0.832     & 0.848     \\
        &EdgeNAT-S2\textdaggerdbl     &      & 0.837     & 0.852     \\
        &EdgeNAT-S2\textdagger\textdaggerdbl &      & 0.840     & 0.859     \\
        \cline{2-2}
        \cline{4-5}
        &EdgeNAT-S3   &           & 0.827     & 0.844     \\
        &EdgeNAT-S3\textdagger    &      & 0.835     & 0.850     \\
        &EdgeNAT-S3\textdaggerdbl     &      & 0.836     & 0.854     \\
        &EdgeNAT-S3\textdagger\textdaggerdbl &      & 0.843     & 0.860     \\
        \cline{2-2}
        \cline{4-5}
        &EdgeNAT-L   &           & 0.843     & 0.859     \\
        &EdgeNAT-L\textdagger    &      & 0.849     & 0.863     \\
        &EdgeNAT-L\textdaggerdbl     &      & \textbf{\textcolor{blue}{0.855}}     & \textbf{\textcolor{blue}{0.870}}     \\
        &EdgeNAT-L\textdagger\textdaggerdbl &      & \textbf{\textcolor{red}{0.860}}     & \textbf{\textcolor{red}{0.876}}     \\
        \bottomrule
    \end{tabular}
    \end{adjustbox}
    \caption{Comparison with other methods on BSDS500 dataset. \textdagger indicates training with extra PASCAL VOC data, and \textdaggerdbl means the multi-scale input. The best two results are highlighted in \textbf{\textcolor{red}{red}} and \textbf{\textcolor{blue}{blue}}, respectively, and same for other tables.}
    \label{tab:compare_BSDS500_tab}
\end{table}

\begin{table}
    \centering
    \resizebox{\columnwidth}{!}{
    \begin{tabular}{c|l|cc|cc|cc}
        \toprule
        \multicolumn{2}{c|}{\multirow{2}{*}{Method}} & \multicolumn{2}{c}{RGB} & \multicolumn{2}{|c}{HHA} & \multicolumn{2}{|c}{RGB-HHA} \\
        \cline{3-8}
        \multicolumn{2}{c|}{} & ODS & OIS & ODS & OIS & ODS & OIS \\
        \midrule
        \multirow{6}{*}{\rotatebox{90}{Traditional}} &gPb-UCM     & 0.632 & 0.661 & - & - & - & - \\
        &gPb+NG     & 0.687 & 0.716 & - & - & - & - \\
        &SE     & 0.695 & 0.708 & - & - & - & - \\
        &SE+NG+     & 0.706 & 0.734 & - & - & - & - \\
        &OEF     & 0.651 & 0.667 & - & - & - & - \\
        &SemiContour     & 0.680 & 0.700 & - & - & - & - \\
        \midrule
        \multirow{8}{*}{\rotatebox{90}{CNN-based}} &HED     & 0.720 & 0.734 & 0.682 & 0.695 & 0.746 & 0.761 \\
        &COB     & -     & -     & -     & -     & 0.784 & \textbf{\textcolor{blue}{0.805}} \\
        &RCF     & 0.729 & 0.742 & 0.705 & 0.715 & 0.757 & 0.771 \\
        &AMH-Net & 0.744 & 0.758 & \textbf{\textcolor{blue}{0.717}} & 0.729 & 0.771 & 0.786 \\
        &LPCB    & 0.739 & 0.754 & 0.707 & 0.719 & 0.762 & 0.778 \\
        &BDCN    & 0.748 & 0.763 & 0.707 & 0.719 & 0.765 & 0.781 \\
        &PiDiNet & 0.733 & 0.747 & 0.715 & 0.728 & 0.756 & 0.773 \\
        &PEdger  & 0.742 & 0.757 & -     & -     & -     & -     \\
        \midrule
        \multirow{6}{*}{\rotatebox{90}{Transformer}} &EDTER   & \textbf{\textcolor{blue}{0.774}} & \textbf{\textcolor{blue}{0.789}} & 0.703 & 0.718 & 0.780 & 0.797 \\
        \cline{2-8}
        &EdgeNAT-S0 & 0.763 & 0.777 & 0.714 & 0.731 & 0.775 & 0.793 \\
        &EdgeNAT-S1 & 0.769 & 0.783 & 0.714 & 0.731 & 0.780 & 0.795 \\
        &EdgeNAT-S2 & 0.773 & 0.787 & \textbf{\textcolor{blue}{0.717}} & \textbf{\textcolor{blue}{0.734}} & 0.783 & 0.799 \\
        &EdgeNAT-S3 & \textbf{\textcolor{blue}{0.774}} & \textbf{\textcolor{blue}{0.789}} & \textbf{\textcolor{blue}{0.717}} & 0.733 & \textbf{\textcolor{blue}{0.785}} & 0.800 \\
        &EdgeNAT-L & \textbf{\textcolor{red}{0.789}} & \textbf{\textcolor{red}{0.803}} & \textbf{\textcolor{red}{0.726}} & \textbf{\textcolor{red}{0.741}} & \textbf{\textcolor{red}{0.794}} & \textbf{\textcolor{red}{0.808}} \\
        \bottomrule
    \end{tabular}}
    \caption{Quantitative comparisons on NYUDv2. All results are computed with a single scale input.}
    \label{tab:Tab_quantitative_comparisons_on_NYUDv2}
\end{table}

\end{document}